\documentclass[preprint,12pt]{elsarticle}

\usepackage{amssymb}
\usepackage{amsmath}

\usepackage{multirow}
\usepackage[capitalize]{cleveref}
\usepackage{color}
\usepackage{colortbl}
\usepackage{algorithmic}
\usepackage{algorithm}

\crefname{section}{Sec.}{Secs.}
\Crefname{section}{Section}{Sections}
\Crefname{table}{Table}{Tables}
\crefname{table}{Tab.}{Tabs.}
\newcommand{\ie}{\textit{i}.\textit{e}.}
\newcommand{\eg}{\textit{e}.\textit{g}.}

\journal{Neural Networks}

\begin{document}

\begin{frontmatter}



\title{Enhancing Consistency and Mitigating Bias: \\
A Data Replay Approach for Incremental Learning} 

        
\author[addr1]{Chenyang Wang}
\ead{cswcy@hit.edu.cn}
\author[addr1]{Junjun Jiang\corref{mycorrespondingauthor}}
\ead{jiangjunjun@hit.edu.cn}
\author[addr1]{Xingyu Hu}
\ead{huxingyu@hit.edu.cn}
\author[addr1]{Xianming Liu}
\ead{csxm@hit.edu.cn}
\address[addr1]{School of Computer Science and Technology, Harbin Institute of Technology, Harbin 150001, China.}

\author[addr2]{Xiangyang Ji}
\address[addr2]{Department of Automation, Tsinghua University, Beijing 100084, China.}
\ead{xyji@tsinghua.edu.cn.}

\cortext[mycorrespondingauthor]{Corresponding author}


\begin{abstract}
Deep learning systems are prone to catastrophic forgetting when learning from a sequence of tasks, as old data from previous tasks is unavailable when learning a new task. To address this, some methods propose replaying data from previous tasks during new task learning, typically using extra memory to store replay data. However, it is not expected in practice due to memory constraints and data privacy issues. Instead, data-free replay methods invert samples from the classification model. While effective, these methods face inconsistencies between inverted and real training data, overlooked in recent works. To that effect, we propose to measure the data consistency quantitatively by some simplification and assumptions. Using this measurement, we gain insight to develop a novel loss function that reduces inconsistency. Specifically, the loss minimizes the KL divergence between distributions of inverted and real data under a tied multivariate Gaussian assumption, which is simple to implement in continual learning. Additionally, we observe that old class weight norms decrease continually as learning progresses. We analyze the reasons and propose a regularization term to balance class weights, making old class samples more distinguishable. \textcolor{black}{To conclude, we introduce Consistency-enhanced data replay with a Debiased classifier for class incremental learning (CwD).} Extensive experiments on CIFAR-100, Tiny-ImageNet, and ImageNet100 show consistently improved performance of CwD compared to previous approaches.
\end{abstract}




\end{frontmatter}



\section{Introduction}
The common practice of training a deep model in the computer vision community is to collect a big enough dataset first and then train offline based on the collected dataset~\cite{deng2009imagenet,Krizhevsky2009-cifar,he2016deep}. However, this kind of training scheme has some limitations. For example, during deployment, when unexpected events occur or new tasks appear, the model needs retraining on both the old and the newly encountered data, which is costly. What is more, data storage is always limited in practice and can not save all the data when tasks continually come. In addition, saving the encountered data permanently may incur privacy issues~\cite{zhang2022deep}. Taking these problems into account, a model that can learn from a sequence of tasks is in need. However, when training data changes, the knowledge learned before in the deep models will be catastrophically forgotten~\cite{mccloskey1989catastrophic}. In this paper, we mainly focus on the class incremental learning (CIL) setting, where data in different tasks are belonging to different classes and the purpose is to classify samples from all classes without the task information.

To avoid catastrophic forgetting, some researchers proposed revising the old knowledge by replaying the data of the experienced tasks when learning the new task. Some works achieved this goal by keeping an extra data memory with a fixed size and designing specific strategies to manage and learn from the limited data in the memory~\cite{rebuffi2017icarl,hou2019learning,chaudhry2019tiny,douillard2020podnet,liang2024loss}. Unlike them, other methods tried to incrementally train a generator to generate old samples~\cite{kamra2017deep,shin2017continual,zhai2019lifelong} and learn from the generated data, which is not constrained by the number of samples but by the quality of the generation. In addition, they still needed memory to save the generator in the whole incremental learning process. Regarding memory constraints and latent privacy issues in practice, both the two lines of methods are not that applicable.

\begin{figure}
  \centering
  \vspace{0.15cm}
  \includegraphics[width=0.95\linewidth]{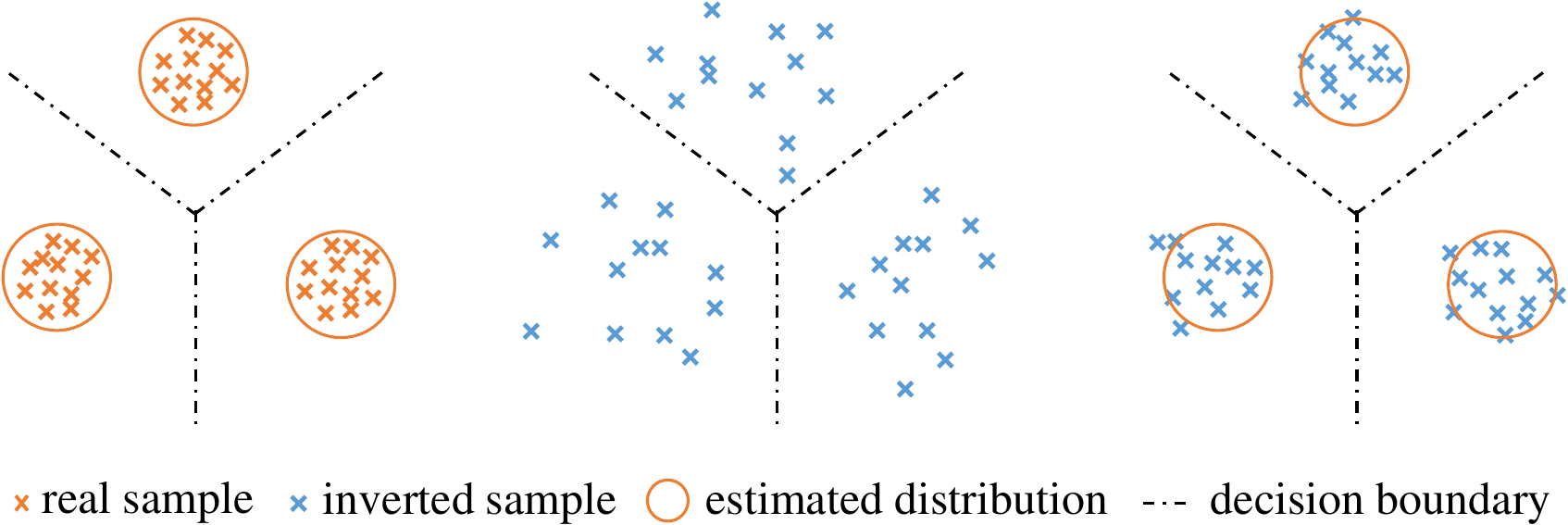}
  \caption{Schematic illustration of data consistency enhancement. Left, the situation of real samples. The distribution is estimated from real samples. Middle, the situation of inverted samples before data consistency enhancement. Right, the situation of inverted samples after consistency enhancement. }
  \label{fig:consistency}
\end{figure}

{\color{black}To overcome limitations, some works~\cite{smith2021always,gao2022r} aim to perform class incremental learning in a data-free manner, called data-free class incremental learning (DFCIL). That is, neither additional data nor models are carried over across tasks.} These works mainly build on the idea of inversion-based data-free knowledge distillation~\cite{yin2020dreaming,nayak2019zero,fang2022up} by inverting samples directly from the classification model. It is natural to introduce the technique into data-free CIL, where at each new learning phase the trained model on experienced tasks is available but the previous data are inaccessible. As for these data-replay-based methods, the memory constraint problem and latent privacy issues are properly sidestepped. However, they still face an inversion quality problem which is severe because of the lack of data. Thus, existing methods have made great efforts on how to distill from these inverted samples and achieve SOTA performance.

Despite their success, the inversion quality problem further causes data inconsistency between synthetic and real samples, which is left unsolved and ignored by existing methods. Unlike them, we focus on the inversion stage and consider how to narrow the distribution gap, which is orthogonal to previous works. The schematic illustration of our motivation can be found in \Cref{fig:consistency}. We argue that efforts in the inversion stage will help the overall performance. In the ideal case, the distribution of synthetic data is the same as the real data and the CIL problem degenerates into an offline supervised learning problem, which is much easier to solve. To that effect, we propose to quantitatively measure the data consistency by some simplification and assumptions. Specifically, we model the data distribution in the representation space of the penultimate layer under a tied multivariate Gaussian assumption, which inspires a new loss term by aligning the estimated distributions of old real and synthetic samples accordingly. It is natural to implement in the incremental learning setting: we estimate the parameters of the distributions at the end of the old task and when the new task comes, we regularize the synthetic samples to have similar statistics as the estimated ones to enhance consistency. 

On the other hand, we observe an unexpected phenomenon during incremental learning, where the norm of weight vectors belonging to old classes is lower than that belonging to new classes. It is similar to the observation in \cite{zhao2020maintaining}, but the underlying reasons are different. In \cite{zhao2020maintaining}, the class imbalance between replay data and new data contributed to the biased class vectors. However, in inversion-based methods, the different learning strategies and different properties of synthetic data and real data take the major responsibility. Unlike \cite{zhao2020maintaining} where a post-processing refinement of class vectors is introduced, we propose to add a regularization term to align the unbiased class weights. In this way, the network and weight vectors are optimized together in an unbiased and matching manner, which is more friendly to the inversion stage.

The main contributions of this work are summarized as follows:
\begin{itemize}
\item We quantitatively measure the data consistency of synthetic and real old data with some simplification, which inspires a novel loss term for enhancing the data consistency.
\item We analyze the underlying reasons for the biased class vectors when learning from both synthetic data and new data and put forward a simple regularization term that is friendly to the inversion stage.
\item Through extensive experiments on different datasets, we show that our method can be combined with different baselines and achieve SOTA performance. 
\end{itemize}

{\color{black}The structure of the paper is as follows. In \Cref{sec:Related}, we discuss related work, including class-incremental learning methods and consistency measurement. \Cref{sec:Preliminary} outlines the problem setting and DFCIL baselines. \Cref{sec:Method} describes the proposed framework, focusing on consistency-enhanced replay and the debiased classifier. \Cref{sec:Experiments} presents experimental comparisons, ablation studies, and further analysis. In \Cref{sec:lim}, we discuss the potential limitations of our approach, and finally, \Cref{sec:Conclusion} summarizes our findings.}

\section{Related Work}
\label{sec:Related}

\subsection{Class Incremental Learning}
There is a rich body of literature on class incremental learning (CIL), with various methods addressing the catastrophic forgetting problem from different perspectives. Some works~\cite{maltoni2019continuous,lee2020neural,yan2021dynamically,wu2021incremental,wu2022class,douillard2022dytox,pomponi2024cascaded} proposed incrementally expanding the network to accommodate new classes from different tasks. These methods allow the modules to function collectively during the testing stage to classify all previously seen classes. Another effective strategy to alleviate catastrophic forgetting is to regularize the optimization of the deep model when learning new tasks. Works in this area aimed to preserve crucial information from past tasks~\cite{hersche2022constrained,shi2021overcoming,akyurek2021subspace,ebrahimi2019uncertainty} or to imitate the oracle model~\cite{zhu2023imitating}. Additionally, some research enhanced the performance of CIL by leveraging extra resources, including unlabeled data~\cite{tang2022learning,lee2019overcoming}, pretrained models~\cite{wu2022class}, and auxiliary tasks~\cite{mazumder2023leveraging}.

The methods~\cite{rebuffi2017icarl,castro2018end,hou2019learning,chaudhry2019tiny,douillard2020podnet,hu2021distilling,wu2024hyper} closely related to ours are based on knowledge distillation~\cite{hinton2015distilling}, where data representing past tasks are available for replay. For example, iCaRL~\cite{rebuffi2017icarl} performed logits distillation between old and new models using both the stored data and the data from the new task. UCIR~\cite{hou2019learning} distilled features between the old and new models, PODNet~\cite{douillard2020podnet} chose to distill pooled feature maps across multiple layers, and \citet{wu2024hyper} introduced hyper-feature aggregation and relaxed distillation to improve knowledge retention and flexibility of adaptation. Instead of storing data, some works~\cite{shin2017continual} additionally kept a generative model in memory to produce enough samples for distillation. Despite their effectiveness and popularity, all of these methods require a relatively large memory budget to perform well.

\subsection{Data-free Class Incremental Learning}
Data-free class incremental learning refers to CIL without storing either generative models or training data from past tasks. To address this, LwF~\cite{li2017learning} substituted data from the new task for previously encountered data for distillation. Due to the distribution differences between new task data and old task data, LwF's performance was relatively limited. In this regard, \citet{zhu2022self} proposed performing distillation only on a subset of new data samples selected by saved prototypes. \citet{shi2023prototype} further updated class prototypes while improving knowledge transfer between old and new classes. To close the gap between data used for distillation and encountered data, replay-based methods~\cite{yin2020dreaming,smith2021always,gao2022r,pourkeshavarzi2021looking,nayak2019zero} directly inverted data for distillation from the classification model. \cite{yin2020dreaming} was the first attempt but failed on the standard CIL benchmark due to the gap between synthetic and real samples. To facilitate learning, \cite{smith2021always} designed modified cross-entropy training and importance-weighted feature distillation. The subsequent work \cite{gao2022r} split the representation learning and classification refinement stages. During representation learning, \cite{gao2022r} adopted a relation-guided distillation loss to alleviate the conflict between plasticity and stability. Additionally, \cite{pourkeshavarzi2021looking} built upon the inversion stage proposed in \cite{nayak2019zero} and introduced an additional strategy to ensure the quality of inverted samples using a memorized confusion matrix. {\color{black}The major distinction between our work and these approaches is the introduction of an estimation stage after training. By estimating statistics of past tasks' data, we enhance the consistency between inverted data and the original data from previous tasks. Furthermore, we propose a weight alignment regularization term to address the classifier bias caused by the learning strategy and inverted samples, which is first reported in the literature.}

\subsection{Data Consistency Measurement}
GAN~\cite{goodfellow2020generative} loss is a widely used and effective loss to generate samples that are indistinguishable from target samples. However, it is not suitable in our setting where no target data is available. Domain alignment methods~\cite{muandet2013domain,motiian2017unified,li2018domain} in domain generalization are also related to the measurement of data consistency. But unlike generating consistent data in our setting, domain alignment methods align data of different domains in the feature space with the help of additional models and/or data. The setting most relative to our work is the inference-time OOD detection~\cite{lee2018simple,morteza2022provable,liu2020energy,sun2021react}, where a pretrained classification model is frozen and training data is accessible (just like the end of the task in incremental learning). Our work builds on the assumptions proposed in \cite{lee2018simple} and proposes to align the statistical parameters in the feature space to enhance data consistency.

\section{Preliminaries}
\label{sec:Preliminary}
\subsection{Problem Settings}
The purpose of class incremental learning is to learn a model that can tell the classes of all encountered training samples that come as a sequence of tasks. We formulate it as follows. Suppose $\mathcal{T}_1, \mathcal{T}_2, \ldots, \mathcal{T}_{N}$ are the $N$ tasks to be learned sequentially. \textcolor{black}{The label sets are disjoint among different tasks: $\forall i,j, i\neq j, \mathcal{C}_i\cap \mathcal{C}_j= \varnothing$.} After learning $\mathcal{T}_1, \mathcal{T}_2, \ldots, \mathcal{T}_{N}$ (denoted as $\mathcal{T}_{1:N}$), the classification model is supposed to classify samples from all seen classes $\mathcal{C}_1 \cup \mathcal{C}_2 \cup \ldots \cup \mathcal{C}_{N}$ (denoted as $\mathcal{C}_{1:N}$). On top of that, at task $\mathcal{T}_i$ in our setting, samples belonging to classes in $\mathcal{C}_i$ are sufficient while no samples belonging to past tasks $\mathcal{T}_{1:i-1}$ are accessible. The classification model, denoted as $f(\cdot)$, consists of a feature extractor $h(\cdot)$, which follows a deep neural network architecture, and a classifier $g(\cdot)$, which represents the final fully connected layer, such that $f(\cdot) = g(h(\cdot))$.

{\color{black}\subsection{Inversion-based DFCIL Approaches}
In this section, we describe the pipeline of inversion-based DFCIL approaches and the specific baseline on which our work is based. Typically, inversion-based methods consist of two main stages: the inversion stage and the training stage. In the DFCIL setting, due to the lack of previous data, the inversion-based method begins with the inversion stage, which directly optimizes samples or trains a generator based only on the old classification model. These optimized samples or the trained generator act as a surrogate for the real old data. Given that the classification model is traditionally designed to classify samples, this inverse data generation process based on it is termed inversion, and samples created in this way are referred to as inverted samples. After the inversion stage, the new task data and inverted data (either offline optimized samples in the inversion stage or online generated data by the trained generator) are used to train the new classification model for all encountered tasks, which forms the training stage.

In chronological order, we briefly introduce representative inversion-based works from the literature. DeepInversion~\cite{yin2020dreaming} was the initial attempt, where samples were directly optimized (treating pixels as parameters) during the inversion stage, and the standard cross-entropy (CE) loss was adopted for these inverted samples. ABD~\cite{smith2021always} extended DeepInversion by making two key modifications: training a generator instead of samples during the inversion stage for efficiency and adopting importance-weighted feature distillation to overcome the problem caused by the gap between the inverted and real samples. R-DFCIL~\cite{gao2022r} retained the same inversion stage as ABD but introduced a relation-guided distillation loss during the training stage to better balance plasticity and stability.

In this paper, we combine the inversion stage proposed in ABD and the training stage proposed in R-DFCIL as our baseline method due to its SOTA performance. To be noticed, the inversion stage of ABD used the same objective function as DeepInversion with different optimizable objects. As a result, the inversion stage is still called DeepInversion. 

The first component of DeepInversion is the cross-entropy loss based on the assumption that inverted samples can be classified correctly by the old model. Let us denote the generator to be trained as $G(\cdot)$ and the old model as $f_{i-1}(\cdot)=g_{1:i-1}(h_{i-1}(\cdot))$. Each online inverted sample is generated as follows: $\hat{x}=G(n), \hat{y}=\mathop{\mathrm{argmax}}\limits_{k}f_{i-1}(\hat{x})_k$, where $n$ follows low-dimensional Gaussian distribution $\mathcal{N}$. Then the cross-entropy loss is given as follows:
\begin{equation}
\label{eq:ce}
  \mathcal{L}_{ce} = \mathop{\mathbb{E}}\limits_{n\sim \mathcal{N}}[\ell_{ce}(f_{i-1}(G(n)), \mathop{\mathrm{argmax}}\limits_{k}f_{i-1}(G(n))_k, \tau)],
\end{equation}
where $\tau$ is the temperature.} Besides, the statistics alignment loss is proposed to align the Batch Normalization (BN) statistics. Each BN layer stores the running mean and variance of features, which can be directly utilized as the prior knowledge to regularize the inversion stage as follows:
\begin{equation}
\label{eq:stat}
    \mathcal{L}_{stat}=\sum_{l} \mathcal{D}_{KL}(\mathcal N(\boldsymbol{\sigma_l}, \boldsymbol{\mu_l}), \mathcal N(\boldsymbol{\hat{\sigma}_l}, \boldsymbol{\hat{\mu}_l})),
\end{equation}
where $l$ indicates the $l$-th BN layer in the model, $\boldsymbol{\sigma_l}, \boldsymbol{\mu_l}$ denotes the mean and variance of the $i$-th BN layer stored in the old model, $\boldsymbol{\hat{\sigma}_l}, \boldsymbol{\hat{\mu}_l}$ denotes the mean and variance estimated on the inverted data, and $\mathcal{D}_{KL}$ denotes the KL divergence. {\color{black}Furthermore, to ensure class balance in the generated data, a class diversity loss is adopted in the baseline as follows:
\begin{equation}
\label{eq:div}
    \mathcal{L}_{div}=H\left(\mathop{\mathbb{E}}\limits_{n\sim \mathcal{N}}\left[{\rm Softmax}(f_{i-1}(G(n)))\right]\right),
\end{equation}
where $H$ denotes the information entropy. The overall loss for training $G(\cdot)$ in the inversion stage is:}
\begin{equation}
    \mathcal{L}_{I\_base}=\mathcal{L}_{ce}+\lambda_{1}\mathcal{L}_{stat}+
    \lambda_{2}\mathcal{L}_{div},
\end{equation}
where $\lambda_1,\lambda_2$ are hyperparameters controlling the scales of the losses. {\color{black}It is worth noting that unlike generative data replay~\cite{shin2017continual}, the generator used here is inverted from the old classification model $f_{i-1}$ in the inversion stage of the $i$-th task and can be immediately discarded after the training stage of the $i$-th task, as it is not stored for future use. In other words, a generator is randomly initialized and trained every time a new task comes.}

{\color{black} During the training stage, we adopted R-DFCIL~\cite{gao2022r} as our baseline. In R-DFCIL, hard knowledge distillation (HKD) was used to distill information from old tasks, local cross-entropy loss (LCE) was employed for learning from current task samples, and relational knowledge distillation (RKD) loss created a connection between old and new models when training with new samples. Specifically, HKD is applied on synthetic samples to prevent model changes as follows:
\begin{equation}
\label{eq:hkd}
  \mathcal{L}_{hkd} = \frac{1}{|\mathcal{C}_{1:i-1}|} \mathop{\mathbb{E}}\limits_{n\sim \mathcal{N}}||f_{i-1}(G(n))-[f_{i}(G(n))]_{1:|\mathcal{C}_{1:i-1}|}||_1,
\end{equation}
where $[f_{i}(G(n))]_{1:|\mathcal{C}_{1:i-1}|}$ denotes the sub-vector of $f_{i}(G(n))$ with dimensions from 1 to $|\mathcal{C}_{1:i-1}|$. Then, a cross-entropy loss is applied locally to the current task's samples (denoted as $X^i$) as follows:
\begin{equation}
\label{eq:lce}
  \mathcal{L}_{lce} = \frac{1}{|X^i|} \sum_{(x,y)\in X^i}\ell_{ce}(g_i(h_{i}(x)), y, \tau),
\end{equation}
where $g_i(h_{i}(x))$ means applying Softmax on the logits of classes belonging to the current $i$-th task. We omit the introduction of the RKD loss for its complexity and its limited relevance to the focus of our work. We denote the baseline training loss $\mathcal{L}_{T\_base}$, which consists of these three terms, as follows:
\begin{equation}
    \mathcal{L}_{T\_base}=\lambda_{3}\mathcal{L}_{hkd} + \lambda_{4}\mathcal{L}_{lce} + \lambda_{5}\mathcal{L}_{rkd},
\end{equation}
where $\lambda_3, \lambda_4, \lambda_5$ are corresponding coefficients.} Notably, our contributions are orthogonal to previous works' that relate to the distillation loss in the training stage and can be combined with any of them (see details in \Cref{sec:performance}). When the choice of the baseline method changes, $\mathcal{L}_{T\_base}$ will change accordingly.

\section{Method}
\label{sec:Method}
\subsection{Overall Framework}
\begin{figure}[t]
  \centering
  \includegraphics[width=0.95\linewidth]{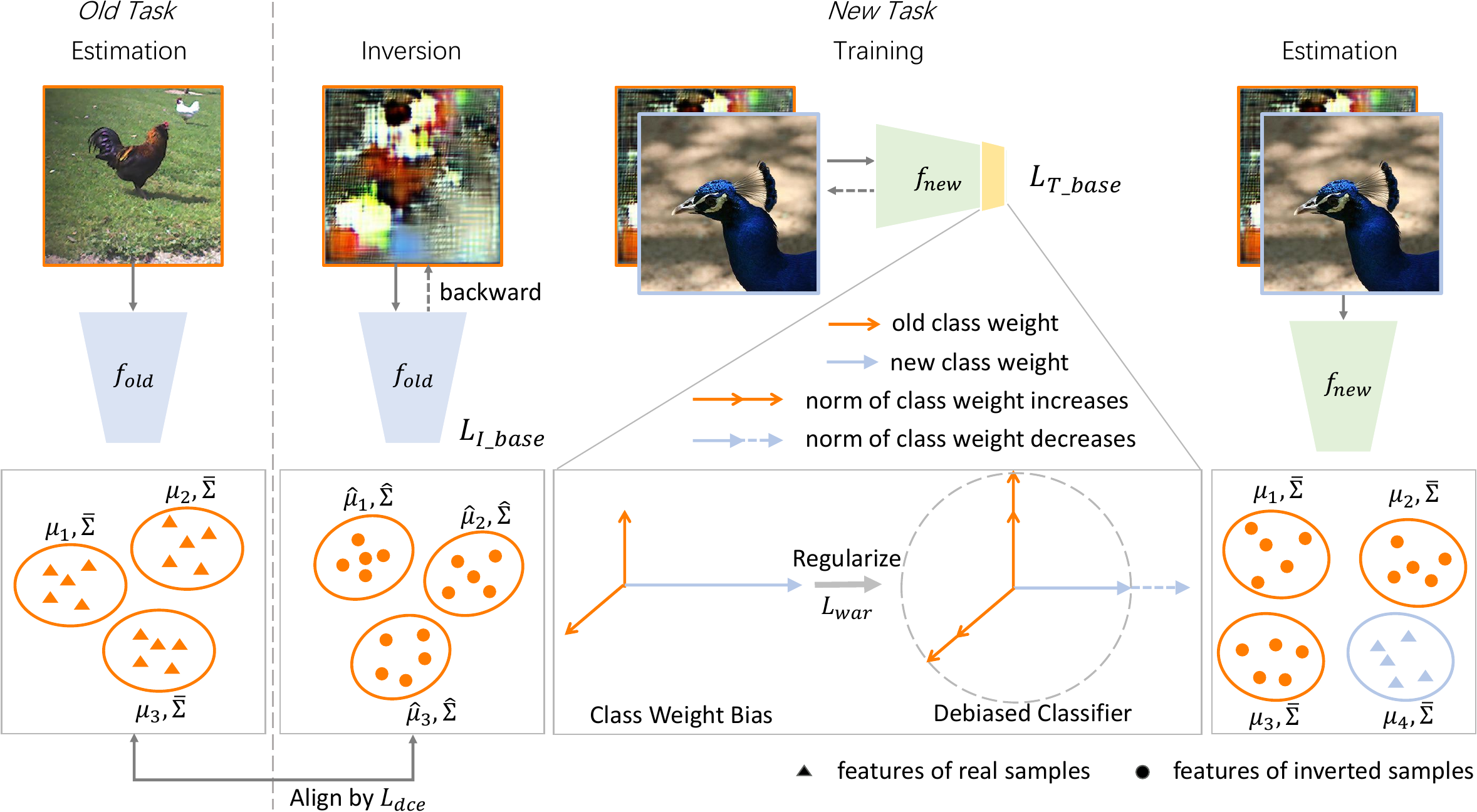}
  \caption{An overview of our proposed CwD framework. Inversion: when a new task comes, we first invert samples from the old model with the help of statistical parameters in the old task. Data consistency enhancement loss $L_{dce}$ is applied in this stage. Training: we use the inverted data and real new data to train a new model. During the training stage, we regularize the class weights to be unbiased by weight alignment regularization loss $L_{war}$. Estimation: when training is over, we estimate the statistical parameters of all classes by the new model.}
  \label{fig:framework}
\end{figure}

The challenge of the incremental learning setting is how to preserve the knowledge of past tasks when past data are inaccessible. To cope with it, we propose the Consistency enhanced data replay with Debiased classifier for class incremental learning (CwD). {\color{black}CwD introduces an additional estimation stage to capture the distribution of old data for better approximation in the next inversion stage. As a result, CwD framework includes one more stage within each task compared to previous methods.} (1) Inversion. The purpose of the inversion stage is to synthesize samples as a substitute for past data. In this stage, we regularize the statistics of synthetic samples to be the same as the past samples with the help of the old model and the statistical parameters from the last task. Our data consistency enhancement (DCE) loss is applied in this stage. (2) Training. In the training stage, a new model inherited from the old model is trained to classify data from all tasks. During the training stage, we propose a weight alignment regularization (WAR) loss to debias the class weights. (3) Estimation. When training is finished, statistical parameters characterizing classes of all encountered tasks will be estimated and memorized for the following task. To be noticed, for the first task, only training and estimation stages are in need and only real samples are involved in the training stage. The overall framework is shown in \Cref{fig:framework}. 

\subsection{Consistency Enhanced Data Replay}
\subsubsection{Estimate Data Consistency} 
\label{sec:estimate}
Our goal is to enhance the consistency between the real and synthetically replayed samples in the inversion stage to facilitate the following training stage of the new task. To begin with, we try to measure the consistency quantitatively. Denote the probability of real samples as $p(x)$ and synthetic samples as $q(x)$, it is natural to measure the consistency of the two distributions by the KL divergence: $D_{KL} = \mathbb E_p\log\frac{p(x)}{q(x)}$. However, directly computing $D_{KL}$ is infeasible in practice due to the high dimension of the image space. What is more, the metrics about pixel differences lack sufficient semantic interpretation. Instead, we pay attention to the output features of the penultimate layer (\ie, $\boldsymbol{z}=h(x)$, the input of the classifier $g(\cdot)$), because it has a relatively low dimension and directly and evidently affects the final classification performance as also discussed in the literature of CIL~\cite{joseph2022energy,caccia2022new} and more~\cite{morteza2022provable,zhou2022learning}.

{\color{black}There are two ways to estimate $p(\boldsymbol{z})$ and $q(\boldsymbol{z})$ in the literature: parametric and non-parametric estimation. Parametric estimation assumes that the data follows a specific distribution or model while non-parametric estimation makes no explicit assumptions about the data distribution. Because non-parametric estimation is inefficient in computing and is not differential, it incurs difficulties when adopting it as part of the objective function in the inversion stage. As a result, we adopt parametric estimation by assuming the class-conditional distribution in the feature space follows a tied multivariate Gaussian distribution.} It is reasonable for the theoretical connection between Gaussian discriminant analysis and the SoftMax classifier and the empirical results detailed in ~\cite{lee2018simple}. To be specific, given any class $k$, denote the feature set of class $k$ as $Z_k$ satisfying: $Z_k\sim\mathcal{N}(\boldsymbol{u_k}, \boldsymbol{\bar{\Sigma}})$, where $\boldsymbol{u_k}$ is the mean vector and $\boldsymbol{\bar{\Sigma}}$ is the tied covariance matrix. The parameters can be estimated easily: $\boldsymbol{u_k} = \frac{1}{|Z_k|}\sum_{\boldsymbol{z}\in Z_k}\boldsymbol{z}$, $\boldsymbol{\bar{\Sigma}}=\frac{1}{|Z|}\left(\sum_k{\sum_{\boldsymbol{z}\in Z_k}\left(\boldsymbol{z}-\boldsymbol{u_k}\right)\left(\boldsymbol{z}-\boldsymbol{u_k}\right)^T}\right)$, where $Z = \bigcup\limits_k Z_k$. In addition, $p(k)$ can be estimated by $|Z_k|/|Z|$. $p(\boldsymbol{z})$ is then given as follows:
\begin{equation}
    p(\boldsymbol{z}) = \sum_k p(k)\mathcal{N}(\boldsymbol{u_k}, \boldsymbol{\bar{\Sigma}}).
\end{equation}
$q(\boldsymbol{z})$ can be estimated similarly is spite that $\hat{Z}$ is the feature set of synthetic samples. Because there is no analytical form of $D_{KL}$ between the mixtures of Gaussians, we adopt the Monte-Carlo approximation method.

\subsubsection{Data Consistency Enhancement Loss}
In this section, we give the details of $L_{dce}$, which intends to align the distributions of real and synthetic data under the aforementioned multivariate Gaussian assumption. Formally, given any class $k\in\mathcal{C}_{1:N}$ and its data $X_k$, denote the feature vectors of $X_k$ as $h(X_k)$ and $h(X_k)$ satisfies: $h(X_k)\sim\mathcal{N}(\boldsymbol{u_k}, \boldsymbol{\bar{\Sigma}})$. In the incremental learning setting, we estimate $\boldsymbol{u_1}, \boldsymbol{u_2}, \dots, \boldsymbol{u_{|\mathcal{C}_{1:i}|}}$ and $\boldsymbol{\bar{\Sigma}}$ at the end of each task $\mathcal{T}_i$ with $h_{i}(\cdot)$ as follows:
\begin{equation}
\label{eq:uk}
    \boldsymbol{u_k} = 
    \begin{cases}
        \frac{1}{|\hat{X}_k^{i-1}|}\sum\limits_{\hat{x}\in\hat{X}_k^{i-1}}h_{i}(\hat{x}), &\text{if $k\notin \mathcal{C}_i$,}\\
        \frac{1}{|X_k|}\sum\limits_{x\in X_k}h_{i}(x), &\text{if $k\in\mathcal{C}_i$,}
    \end{cases}
\end{equation}
\begin{align}
\label{eq:sigma}
    \boldsymbol{\bar{\Sigma}}=&\frac{1}{|\hat{X}^{i-1}\cup X^i|}\Big(\sum_{k\notin\mathcal{C}_i}{\sum_{\hat{x}\in \hat{X}_k^{i-1}}\left(h_{i}(\hat{x})-\boldsymbol{u_k}\right)\left(h_{i}(\hat{x})-\boldsymbol{u_k}\right)^T}\nonumber \\
    &+\sum_{k\in\mathcal{C}_i}{\sum_{x\in X_k}\left(h_{i}(x)-\boldsymbol{u_k}\right)\left(h_{i}(x)-\boldsymbol{u_k}\right)^T}\Big),
\end{align}
where $\hat{X}^{i-1}$ denotes the data inverted from $f_{i-1}$ and $\hat{X}_k^{i-1}$ denotes the inverted data of class $k$. {\color{black}To precisely approximate the distribution of the old data, we try to collect as many samples as possible. For real data $X^i$, we spend an extra epoch collecting all the samples accessible in the current task $i$. For inverted samples $X^{i-1}$, the number of collected samples is adjusted to be $|\mathcal{C}_{1:i-1}|/|\mathcal{C}_i|$ times the number of real samples to maintain balance.} Then, to align the synthetic data $\hat{X}^{i}$ in task $\mathcal{T}_{i+1}$ and data $\hat{X}^{i-1}\cup X^i$ in task $\mathcal{T}_i$, we estimate corresponding parameters of $\hat{X}^{i}$ as follows:
\begin{equation}
    \boldsymbol{\hat{u}_k} = 
    \frac{1}{|B_k|}\sum\limits_{\hat{x}\in\hat{X}_k^{i}\cap B}h_{i}(\hat{x})
\end{equation}
\begin{equation}
    \boldsymbol{\hat{\Sigma}}=\frac{1}{|B|}\sum_{k}{\sum_{\hat{x}\in \hat{X}_k^{i}\cap B}\left(h_{i}(\hat{x})-\boldsymbol{\hat{u}_k}\right)\left(h_{i}(\hat{x})-\boldsymbol{\hat{u}_k}\right)^T},
\end{equation}
where $|B_k|$ is the number of inverted samples belonging to class $k$ in batch $B$ and $|B|$ is the batch size. To enhance the consistency between inverted and real data, we directly align the parameters as follows:
\begin{equation}
\label{eq:dce}
    \mathcal{L}_{dce} = \sum_{k}||\boldsymbol{\hat{u}_k}-\boldsymbol{u_k}||_2+||\boldsymbol{\hat{\Sigma}}- \boldsymbol{\bar{\Sigma}}||_F.
\end{equation}
In ideal situation, we guarantee $p(\boldsymbol{z}|k)=q(\boldsymbol{z}|k)$ by \Cref{eq:dce} and $p(k)=q(k)$ by \Cref{eq:div}. Thus, we align $p(\boldsymbol{z})$ and $q(\boldsymbol{z})$. The overall loss for our consistency-enhanced data replay training is:
\begin{equation}
\label{eq:inv}
    \mathcal{L}_{inv}=\mathcal{L}_{I\_base}+\lambda_{dce}\mathcal{L}_{dce}.
\end{equation}
The proposed DCE loss can be naturally implemented in the context of the DFCIL setting. Specifically, it is viable to estimate the old class distributions with all training samples at the end of the task. And the estimated statistical parameters can be combined with the old model seamlessly to enhance the data consistency in the inversion stage of the new task.

\subsection{Debiased Classifier}
\label{sec:war}
\begin{figure}
  \centering
  \hspace{-0.5cm}\includegraphics[width=0.85\linewidth]{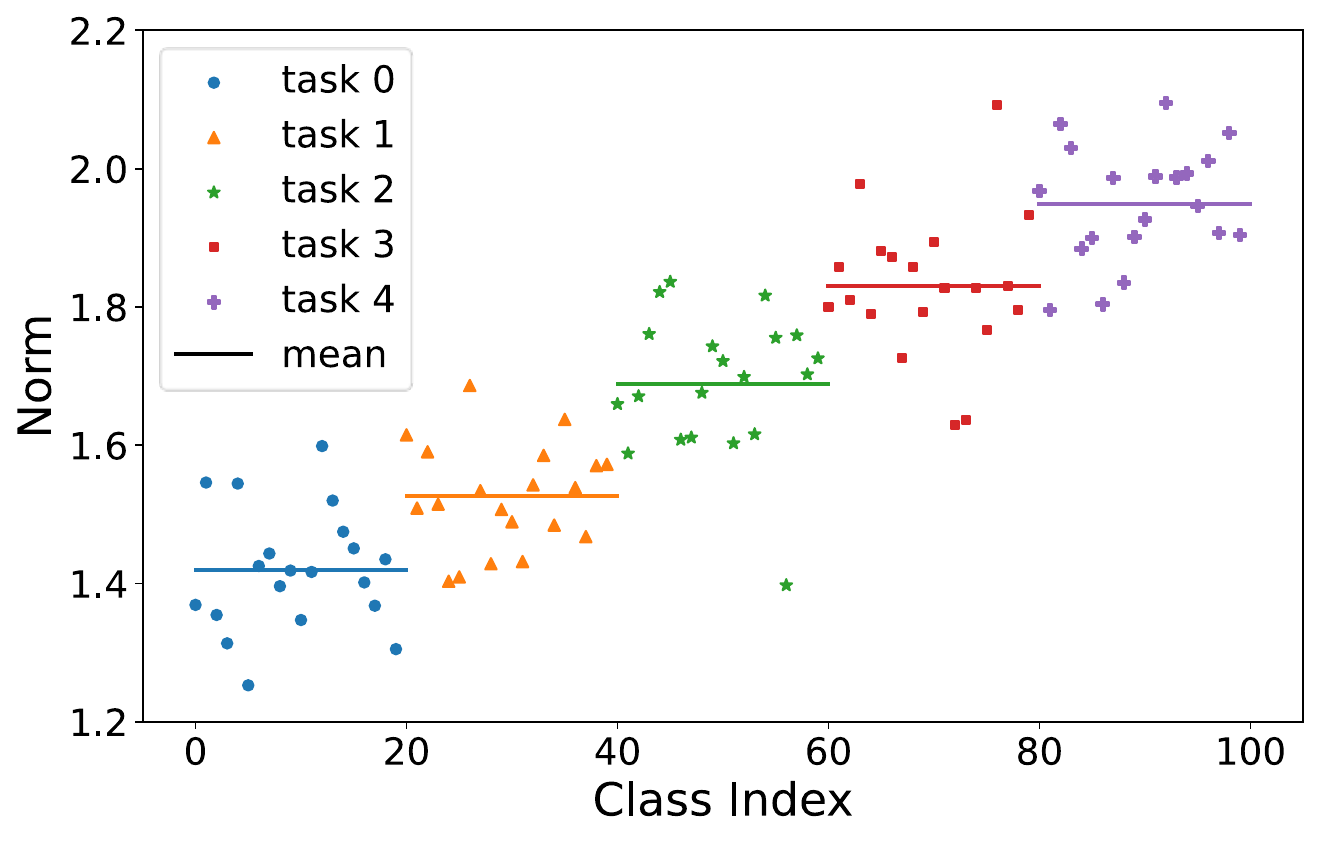}
  \caption{The norms of class weights in the standard 5-task setting.}
  \label{fig:5tasks}
\end{figure}

By treating real and synthetic samples differently, the works of \cite{smith2021always,gao2022r} successfully circumvented the domain classification problem and achieved high performance. However, based on our observation, it incurs a class weight bias between old and new classes: the norm of weights of old classes is smaller than that of new classes and the gap continually widens as learning progresses. Denote the parameters in $g(\cdot)$ as $\boldsymbol{W}\in \mathbb R^{K\times d}$, where $d$ is the dimension of feature $h(x)$ and $K$ is the number of classes. $k$-th row vector in $\boldsymbol{W}$ is denoted as $\boldsymbol{w_k}$, which is the weight vector corresponding to class $k$. We plot the weight norm versus the class order of tasks in \Cref{fig:5tasks}. {\color{black}Analysis of the underlying reasons can be found in \Cref{sec:bias}.}

To correct the bias, we try to align the weights of old and new classes. Though a post-processing alignment strategy is provided in \cite{zhao2020maintaining}, it is not that suitable for data consistency enhancement and will still cause bias among tasks as shown in \Cref{sec:debias}. We prefer to replay data through an unbiased trained model to maximize the satisfaction of the Gaussian assumption. If we align the weights after training, the feature extractor will be biased and the tied covariance may be a biased estimation. So we regularize it during training. Denoting the norm of $||\boldsymbol{w_k}||_2$ as $n_k$, the mean norm of old and new class weights as $n_{old}, n_{new}$, the regularization is as follows: 
\begin{equation}
  \mathcal{L}_{war} = \frac{1}{|C_{0:i}|}\big(\sum_{k\notin\mathcal{C}_i}\left|n_k-n_{new}\right|+\sum_{k\in\mathcal{C}_i}\left|n_k-n_{old}\right|\big),
\end{equation}
As a result, the overall loss for the training stage is:
\begin{equation}
\label{eq:train}
  \mathcal{L}_{train} = \mathcal{L}_{T\_base} + \lambda_{war}\mathcal{L}_{war}.
\end{equation}

{\color{black}Finally, the algorithm that includes the complete incremental learning process is given in Algorithm 1.

\begin{algorithm}[H]
\caption{CwD}
\label{alg:resmooth}
\begin{algorithmic}[1]
\color{black}
\REQUIRE Data sequence $\{X^i\}_{i=1}^N$, Model $f_1(\cdot)$
\ENSURE Model $f_N(\cdot)$
\STATE Randomly initialize $f_1(\cdot)=g_1(h_1(\cdot))$
\FOR {task $i\in\{1,2,\cdots,N\}$}
\IF {$i=1$}
\FOR {$B^{new}\sim X^1$}
\STATE Computer loss $\mathcal{L}_{lce}$ on batch $B^{new}$
\STATE Update $f_1(\cdot)$ with $\mathcal{L}_{lce}$
\ENDFOR \hfill // training stage of the first task
\ELSE
\STATE Randomly initialize $G(\cdot)$
\FOR {$B^{noise}\sim \mathcal{N}$}
\STATE Compute $\mathcal{L}_{inv}$ on $B^{noise}$ with $f_{i-1}(\cdot)$, $\{\boldsymbol{u_k}\}$, and $\boldsymbol{\bar{\Sigma}}$ as in \cref{eq:inv}
\STATE Update $G(\cdot)$ with $\mathcal{L}_{inv}$
\ENDFOR \hfill // inversion stage
\STATE Initialize $h_i(\cdot)=h_{i-1}(\cdot)$ and randomly initialize $g_i(\cdot)$. Combine $g_i(\cdot)$ and $g_{1:i-1}(\cdot)$ as $g_{1:i}(\cdot)$. Initialize $f_i(\cdot)=h_i(g_{1:i}(\cdot))$.
\FOR {$B^{new}\sim X^i$, $B^{noise}\sim \mathcal{N}$}
\STATE Compute $\mathcal{L}_{train}$ on $B^{new}$ and $B^{noise}$ with $f_{i-1}(\cdot)$, $G(\cdot)$ as in \cref{eq:train}
\STATE Update $f_i(\cdot)$ with $\mathcal{L}_{train}$
\ENDFOR \hfill // training stage
\ENDIF
\STATE Initialize the inverted set: $\hat{X}=\varnothing$
\FOR {$B^{noise}\sim \mathcal{N}$}
\STATE Generate a batch: $B^{inv}=G(B^{noise})$
\STATE Append inverted set: $\hat{X}=\hat{X}\cup B^{inv}$
\ENDFOR
\STATE Estimate $\{\boldsymbol{u_k}\}$ and $\boldsymbol{\bar{\Sigma}}$ as in \cref{eq:uk} and \cref{eq:sigma} \hfill // estimation stage
\ENDFOR
\RETURN $f_N\mathcal(\cdot)$
\end{algorithmic}
\end{algorithm}}

\section{Experiments}
\label{sec:Experiments}
{\color{black}In this section, we present experiments to evaluate the effectiveness of our proposed CwD framework. First, we compare CwD with other state-of-the-art methods on several class-incremental learning benchmarks to demonstrate its superiority. Next, we evaluate the performance of WAR regularization against other debiasing techniques. We also conduct an ablation study to assess the impact of key components within the CwD framework. Furthermore, we analyze hyperparameter, data consistency, weight bias, and computational overhead.}

\subsection{Experimental Settings}
We compare the performance of methods on three vision classification benchmark datasets: CIFAR-100~\cite{Krizhevsky2009-cifar}, Tiny-ImageNet~\cite{le2015tiny}, ImageNet-100~\cite{deng2009imagenet}. All three datasets are composed of natural images but differ in the scales of the datasets and sizes of contained images. CIFAR-100~\cite{Krizhevsky2009-cifar} consists of 60K images where each class has 500 images for the training and 100 images for the test. CIFAR-100 is a relatively small dataset and the image size of CIFAR-100 is 32$\times$32. Tiny-ImageNet~\cite{le2015tiny} is a medium-sized dataset, which contains 100K images of 200 classes. The image size of Tiny-ImageNet is larger than that of CIFAR-100 and up to 64$\times$64. ImageNet~\cite{deng2009imagenet} is a large visual dataset consisting of natural images in high resolution. We use the ImageNet-100 subset for efficiency, which has 100 classes and 1.3K training images and 50 validation images per class. The three datasets cover images from different sizes and all have enough classes and samples per class for incremental learning. Specifically, we follow prior works~\cite{gao2022r,smith2021always} to equally split the classes into 5, 10, and 20 tasks and learn from them continually. 

The basic training settings of CwD are kept the same as baselines for fair comparison~\cite{smith2021always, gao2022r}. Specifically, the backbone network for CIFAR-100 and Tiny-ImageNet is a 32-layer ResNet~\cite{he2016deep}. And for ImageNet-100, we adopt Resnet18~\cite{he2016deep} as the backbone. At each task of CIFAR-100 or Tiny-ImageNet, we train the model for 200 epochs. The learning rate is set at 0.1 initially and decayed by 10 at epochs 80 and 120. The weight decay is 0.0005 for CIFAR-100 and 0.0002 for Tiny-ImageNet. For ImageNet-100, we train the model for 120 epochs. The learning rate is 0.1 and decayed at epochs 30 and 60, and the weight decay is set at 0.0001. {\color{black}For all datasets, we use the SGD optimizer during the training stage, with a batch size of 128 for CIFAR-100 and Tiny-ImageNet, and 64 for ImageNet-100. The same batch size is used for both real and inverted samples.} For inversion, we use different generative models for different datasets to generate images of the same size as the real ones. We use the Adam optimizer with a constant learning rate 0.001. We train the generative model for 5000 steps when synthesizing images of CIFAR and Tiny-ImageNet and we train for 10000 steps when synthesizing images of ImageNet. {\color{black}The batch size during the inversion stage is kept consistent with that in the training stage.}

We perform CIFAR-100 and Tiny-ImageNet experiments on an RTX 2080Ti GPU while we perform ImageNet-100 experiments on an V100 GPU. For CIFAR-100 and Tiny-ImageNet experiments, we report the mean$\pm$std\% results based on 3 different class orders. For ImageNet-100 experiments, we report the result based on the same class order. The class orders are set as the same in \cite{gao2022r}.

\subsection{Incremental Learning Performance}
\label{sec:performance}

{\color{black}To validate the effectiveness of the proposed CwD method, we compare it with several baselines, categorized based on whether they are data-free. Non-data-free methods require additional memory to store either data samples or a generative model. Note that while these methods are not directly comparable to ours, their results are included to provide a broader comparison and a more comprehensive understanding.} \\
{\bf Deep Generative Replay~\cite{shin2017continual} (DGR):} a generative data replay method, which needs to keep a generative model all the time during the whole incremental learning process. \\
{\color{black}{\bf Experience Replay~\cite{chaudhry2019tiny} (ER):} a classic data replay method that maintains a memory buffer of past data for sampling at the new task. \\
{\bf Loss Decoupling~\cite{liang2024loss} (LODE):} a SOTA method based on loss decoupling, which separates the objectives for distinguishing new/old classes and distinguishing within new classes. \\
In contrast, data-free methods do not store data samples or generative models across tasks. These methods are briefly introduced as follows:} \\
{\bf Learning without Forgetting~\cite{li2017learning} (LwF):} a classic data-free class incremental method. \\
{\bf DeepInversion~\cite{yin2020dreaming}:} a data-free data replay-based class incremental method, which replays data by model inversion without saving a generative model. \\
{\bf Always Be Dreaming~\cite{smith2021always} (ABD):} a strong baseline, which keeps the loss in inversion stage the same as \cite{yin2020dreaming} with different optimizable objects and changes the training stage for better performance. \\
{\bf Relation-Guided Representation Learning~\cite{gao2022r} (R-DFCIL):} the SOTA baseline, which keeps the inversion stage the same as \cite{smith2021always} and further ameliorates the training stage.

To measure the performance, we adopt the last incremental accuracy $A_N$ and the mean incremental accuracy $\bar{A}$ as metrics. $A_N$ denotes the classification accuracy on all seen classes after learning the last task. $\bar{A}=\sum_{i=1}^N A_i$ averages the results of different learning phases and measures the performance of the model through the whole training. Formally, $A_i$ is defined as follows:
\begin{equation}
  A_i = \frac{1}{|X_{test}^{1:i}|}\sum_{(x,y)\in X_{test}^{1:i}}\mathbb{I}(\mathop{\mathrm{argmax}}\limits_{k}f_{i}(x)_k=y),
\end{equation}
where $\mathbb{I}(\cdot)$ is the indicator function.

\begin{table*}[t]
\tiny
\centering
\caption{Performance on CIFAR-100. The dataset is divided into 5, 10 and 20 tasks. The subscript number of ER and LODE indicates the length of the data buffer. * indicates the results reported in \protect\cite{gao2022r}.}
\setlength{\tabcolsep}{1.6mm}{\begin{tabular}{lcccccc}
\hline
\multirow{2}{*}{Method} & \multicolumn{2}{c}{5}                                 & \multicolumn{2}{c}{10}                                & \multicolumn{2}{c}{20}                                \\ \cline{2-7} 
    \rule {0pt}{7pt}    & $A_N$                     & $\bar{A}$                 & $A_N$                     & $\bar{A}$                 & $A_N$                     & $\bar{A}$                 \\ \hline
DGR*                    & 14.40 $\pm$ 0.40          & -                         & 8.10 $\pm$ 0.10           & -                         & 4.10 $\pm$ 0.30           & -                         \\
{\color{black}ER$_{500}$}              & {\color{black}22.21 $\pm$ 0.36}          & {\color{black}45.71 $\pm$ 1.25}          & {\color{black}16.89 $\pm$ 0.73}          & {\color{black}39.98 $\pm$ 1.48}          & {\color{black}14.35 $\pm$ 0.45}          & {\color{black}36.89 $\pm$ 1.34}          \\
{\color{black}ER$_{5120}$}             & {\color{black}50.35 $\pm$ 0.68}          & {\color{black}66.19 $\pm$ 1.65}          & {\color{black}49.63 $\pm$ 0.13}          & {\color{black}66.95 $\pm$ 1.57}          & {\color{black}48.71 $\pm$ 0.18}          & {\color{black}66.62 $\pm$ 1.54}          \\
{\color{black}LODE$_{500}$}            & {\color{black}31.91 $\pm$ 0.53}          & {\color{black}52.92 $\pm$ 1.31}          & {\color{black}27.03 $\pm$ 0.31}          & {\color{black}48.28 $\pm$ 1.40}          & {\color{black}22.54 $\pm$ 0.75}          & {\color{black}44.19 $\pm$ 1.23}          \\
{\color{black}LODE$_{5120}$}           & {\color{black}54.59 $\pm$ 0.64}          & {\color{black}68.63 $\pm$ 1.84}          & {\color{black}51.98 $\pm$ 0.58}          & {\color{black}67.69 $\pm$ 1.79}          & {\color{black}49.22 $\pm$ 0.35}          & {\color{black}66.12 $\pm$ 1.54}          \\ \hline
LwF*                    & 17.00 $\pm$ 0.10          & -                         & 9.20 $\pm$ 0.00           & -                         & 4.70 $\pm$ 0.10           & -                         \\
DeepInversion*          & 18.80 $\pm$ 0.30          & -                         & 10.90 $\pm$ 0.60          & -                         & 5.70 $\pm$ 0.30           & -                         \\
ABD                     & 46.79 $\pm$ 0.21          & 63.16 $\pm$ 1.58          & 37.01 $\pm$ 1.46          & 57.32 $\pm$ 2.32          & 22.14 $\pm$ 0.63          & 44.53 $\pm$ 1.62          \\
R-DFCIL*                & 50.47 $\pm$ 0.43          & 64.85 $\pm$ 1.78          & 42.37 $\pm$ 0.72          & 59.41 $\pm$ 1.76          & 30.75 $\pm$ 0.12          & 48.47 $\pm$ 1.90          \\
R-DFCIL                 & 49.90 $\pm$ 0.23          & 64.78 $\pm$ 2.23          & 42.57 $\pm$ 0.71          & 59.13 $\pm$ 1.70          & 30.35 $\pm$ 0.12          & 47.80 $\pm$ 1.81          \\ \hline
CwD(ABD)                & 50.27 $\pm$ 0.51          & 64.89 $\pm$ 1.24          & 40.01 $\pm$ 0.86          & 58.86 $\pm$ 2.00          & 25.46 $\pm$ 0.40          & 46.85 $\pm$ 1.43          \\
CwD(R-DFCIL)            & \textbf{52.46 $\pm$ 0.35} & \textbf{66.31 $\pm$ 1.46} & \textbf{43.69 $\pm$ 0.57} & \textbf{60.14 $\pm$ 1.87} & \textbf{31.72 $\pm$ 0.22} & \textbf{49.01 $\pm$ 1.81} \\ \hline
\end{tabular}}
\label{tab:cifar}
\end{table*}

\begin{table*}[t]
\tiny
\centering
\caption{Performance on Tiny-ImageNet. The dataset is divided into 5, 10 and 20 tasks. The subscript number of ER and LODE indicates the length of the data buffer. * indicates the results reported in \protect\cite{gao2022r}.}
\setlength{\tabcolsep}{1.7mm}{\begin{tabular}{lcccccc}
\hline
\multirow{2}{*}{Method} & \multicolumn{2}{c}{5}               & \multicolumn{2}{c}{10}              & \multicolumn{2}{c}{20}             \\ \cline{2-7} 
                        \rule {0pt}{7pt} & $A_N$            & $\bar{A}$   & $A_N$            & $\bar{A}$   & $A_N$            & $\bar{A}$   \\ \hline
{\color{black}ER$_{500}$}              & {\color{black}14.58 $\pm$ 0.23}          & {\color{black}32.21 $\pm$ 0.14}          & {\color{black}8.69 $\pm$ 0.25}          & {\color{black}24.91 $\pm$ 0.46}          & {\color{black}5.17 $\pm$ 0.30}          & {\color{black}20.00 $\pm$ 0.49}          \\
{\color{black}ER$_{5120}$}             & {\color{black}25.32 $\pm$ 0.39}          & {\color{black}46.78 $\pm$ 0.46}          & {\color{black}23.44 $\pm$ 0.17}          & {\color{black}41.85 $\pm$ 0.73}          & {\color{black}22.14 $\pm$ 0.42}          & {\color{black}41.86 $\pm$ 0.42}          \\
{\color{black}LODE$_{500}$}            & {\color{black}18.79 $\pm$ 0.09}          & {\color{black}35.89 $\pm$ 0.41}          & {\color{black}14.00 $\pm$ 0.41}          & {\color{black}31.43 $\pm$ 0.33}          & {\color{black}11.08 $\pm$ 0.62}          & {\color{black}27.50 $\pm$ 0.29}          \\
{\color{black}LODE$_{5120}$}           & {\color{black}31.76 $\pm$ 0.49}          & {\color{black}42.65 $\pm$ 0.51}          & {\color{black}27.92 $\pm$ 0.36}          & {\color{black}45.02 $\pm$ 0.67}          & {\color{black}26.20 $\pm$ 0.13}          & {\color{black}43.74 $\pm$ 0.35}          \\ \hline
ABD                     & 30.40 $\pm$ 0.78 & 45.07 $\pm$ 0.78 & 22.50 $\pm$ 0.62 & 40.52 $\pm$ 0.71 & 15.65 $\pm$ 0.95 & 35.00 $\pm$ 0.53 \\
R-DFCIL*                & 35.89 $\pm$ 0.75 & 48.96 $\pm$ 0.40 & 29.58 $\pm$ 0.51 & 44.36 $\pm$ 0.18 & 24.43 $\pm$ 0.82 & 39.34 $\pm$ 0.18 \\
R-DFCIL                 & 35.25 $\pm$ 0.57 & 48.90 $\pm$ 1.03 & 29.96 $\pm$ 0.36 & 44.58 $\pm$ 0.68 & 24.07 $\pm$ 0.28 & 39.06 $\pm$ 0.62 \\ \hline
CwD(ABD)               & 33.04 $\pm$ 0.30 & 46.95 $\pm$ 0.73 & 25.45 $\pm$ 0.48 & 42.56 $\pm$ 1.05 & 16.92 $\pm$ 0.56 & 36.09 $\pm$ 0.63 \\
CwD(R-DFCIL)           & \textbf{36.89 $\pm$ 0.73} & \textbf{49.69 $\pm$ 0.76} & \textbf{30.90 $\pm$ 0.52} & \textbf{45.29 $\pm$ 0.73} & \textbf{24.54 $\pm$ 0.27} & \textbf{39.65 $\pm$ 0.19} \\ \hline
\end{tabular}}
\label{tab:tiny}
\end{table*}

\begin{table}[t]
\centering
\caption{Performance on ImageNet-100. The dataset is divided into 5, 10 and 20 tasks. The subscript number of ER and LODE indicates the length of the data buffer. * indicates the results reported in \protect\cite{gao2022r}.}
\begin{tabular}{lcccccc}
\hline
\multirow{2}{*}{Method} & \multicolumn{2}{c}{5}             & \multicolumn{2}{c}{10}            & \multicolumn{2}{c}{20}            \\ \cline{2-7} 
                        \rule {0pt}{13pt} & $A_N$            & $\bar{A}$   & $A_N$            & $\bar{A}$   & $A_N$            & $\bar{A}$   \\ \hline
{\color{black}ER$_{500}$}              & {\color{black}24.24}          & {\color{black}46.37}          & {\color{black}15.84}          & {\color{black}37.94}          & {\color{black}13.10}          & {\color{black}31.45}          \\
{\color{black}ER$_{5120}$}             & {\color{black}52.14}          & {\color{black}67.86}          & {\color{black}49.16}          & {\color{black}65.82}          & {\color{black}48.76}          & {\color{black}65.68}          \\
{\color{black}LODE$_{500}$}            & {\color{black}35.52}          & {\color{black}55.26}          & {\color{black}26.90}          & {\color{black}47.39}          & {\color{black}22.08}          & {\color{black}40.58}          \\
{\color{black}LODE$_{5120}$}           & {\color{black}58.18}          & {\color{black}71.33}          & {\color{black}53.58}          & {\color{black}68.09}          & {\color{black}52.30}          & {\color{black}66.86}          \\ \hline
ABD                     & 52.04 & 67.00                     & 38.34 & 58.08                     & 21.74 & 44.69                     \\
R-DFCIL*                & 53.10 & 68.15                     & 42.28 & 59.10                     & 30.28 & 47.33                     \\
R-DFCIL                 & 50.90 & 67.72                     & 41.38 & 58.82                     & 27.86 & 45.74                     \\ \hline
CwD(ABD)               & 56.02 & 69.42                     & 41.86 & 60.52                     & 25.88 & 48.10                     \\
CwD(R-DFCIL)           & \textbf{56.80} & \textbf{70.39} & \textbf{46.04} & \textbf{62.66} & \textbf{34.60} & \textbf{49.18} \\ \hline
\end{tabular}
\label{tab:imagenet}
\end{table}

Because CwD focuses on the inversion stage and the regularization of the classifier, which is orthogonal to SOTA works, we combine it with them (R-DFCIL and ABD) for evaluation. To make a fair comparison, we keep the specific training settings of CwD the same as the combined baseline. If not explicitly mentioned, CwD refers to CwD + R-DFCIL in the following. 

We report the results of CIFAR-100 in \Cref{tab:cifar}, the results of Tiny-ImageNet in \Cref{tab:tiny}, and the results of ImageNet-100 in \Cref{tab:imagenet}. From all three tables, we can see that our CwD method can achieve consistent improvements both in $A_N$ and $\bar{A}$ compared to the data-free baselines. Specifically, CwD surpasses the second-best method by 2.5, 1.1, and 1.4 percent in $A_N$ on CIFAR-100. Likely, the absolute gains are 1.6, 0.9, and 0.5 on Tiny-ImageNet, and the improvements achieve 5.9, 4.7, and 6.7 on ImageNet-100. {\color{black}When comparing to non-data-free methods, we observe two notable phenomena. First, when the number of stored samples is relatively small (500), replay-based methods underperform compared to CwD. However, as the number of stored samples increases (5120), these replay-based methods begin to outperform CwD. Second, CwD is more susceptible to forgetting as the number of tasks grows. For instance, CwD achieves comparable results to LODE$_{5120}$ when the number of tasks is 5 in CIFAR-100 and ImageNet-100, but falls significantly behind when the number of tasks increases to 20. We attribute this to the accumulation of forgetting due to the lack of long-term memory.} What is more, we can see that both R-DFCIL and ABD can benefit from the CwD framework consistently, which verifies the applicability of CwD. In addition, the results in $\bar{A}$ are consistent with results in $A_N$, which indicates that CwD can improve performance in the whole learning process.  

\begin{table}[t]
\centering
\caption{Comparisons with other debiasing methods on CIFAR-100. All numbers in the table are last incremental accuracy (mean$\pm$std\%).}
\begin{tabular}{lccc}
\hline
Method & \multicolumn{1}{c}{5} & \multicolumn{1}{c}{10} & \multicolumn{1}{c}{20} \\ \hline
SCE    & 47.86 $\pm$ 0.51            & 36.00 $\pm$ 0.55             & 22.43 $\pm$ 0.76             \\
ACE    & 47.11 $\pm$ 0.32            & 35.13 $\pm$ 0.91             & 21.75 $\pm$ 0.83             \\
SSIL   & 51.52 $\pm$ 0.31            & 42.81 $\pm$ 0.90             & 28.96 $\pm$ 0.11             \\
WA     & 51.08 $\pm$ 0.58            & 42.41 $\pm$ 0.59             & 29.81 $\pm$ 0.32             \\ \hline
CwD    & \textbf{52.46 $\pm$ 0.35} & \textbf{43.69 $\pm$ 0.57} & \textbf{31.72 $\pm$ 0.22}         \\ \hline
\end{tabular}
\label{tab:debias}
\end{table}

\subsection{Comparisons with Other Debiasing Approaches}
\label{sec:debias}

In \Cref{sec:war}, we propose a simple regularization term $L_{war}$ to align the weights. In this section, we compare the proposed strategy with some known approaches in the literature that can help reduce bias. We briefly describe some approaches as follows: \\
{\bf Split Cross Entropy (SCE):} SCE applies two independent local CE losses (see \Cref{eq:lce}) on the replay data and the data from the new task. Local means the Softmax is calculated locally on old classes or new classes. \\
{\bf Asymmetric Cross Entropy~\cite{caccia2022new} (ACE):} ACE applies the local CE loss on the data from the new task and applies the global CE loss on the replay data, which further impedes the improvement of the norms of new classes. \\
{\bf Separated-Softmax for Incremental Learning~\cite{ahn2021ss} (SSIL):} On top of SCE, SSIL adopts task-wise KD loss to preserve knowledge within each task to avoid bias among tasks. \\
{\bf Weight Aligning~\cite{zhao2020maintaining} (WA):} WA is a post-processing alignment method. After training the model, WA aligns the mean norm of old and new classes by multiplying the weight vectors of new classes by a scalar.

The approaches can be divided into two classes: (1) split losses for data of old and new classes or even for data of each task (SCE, ACE, and SSIL), (2) explicit class weight post-processing (WA). For fair comparisons, we first set $\lambda_{war}=0$ to disable $\mathcal{L}_{war}$ and keep the inversion stage same as in CwD. Then, for the first line of approaches, we replace $L_{hkd}$ with respective proposed losses. For WA, we keep both the inversion and training stage the same as CwD and post-process weights after training.

\begin{figure*}
  \centering
  \includegraphics[width=1.0\linewidth]{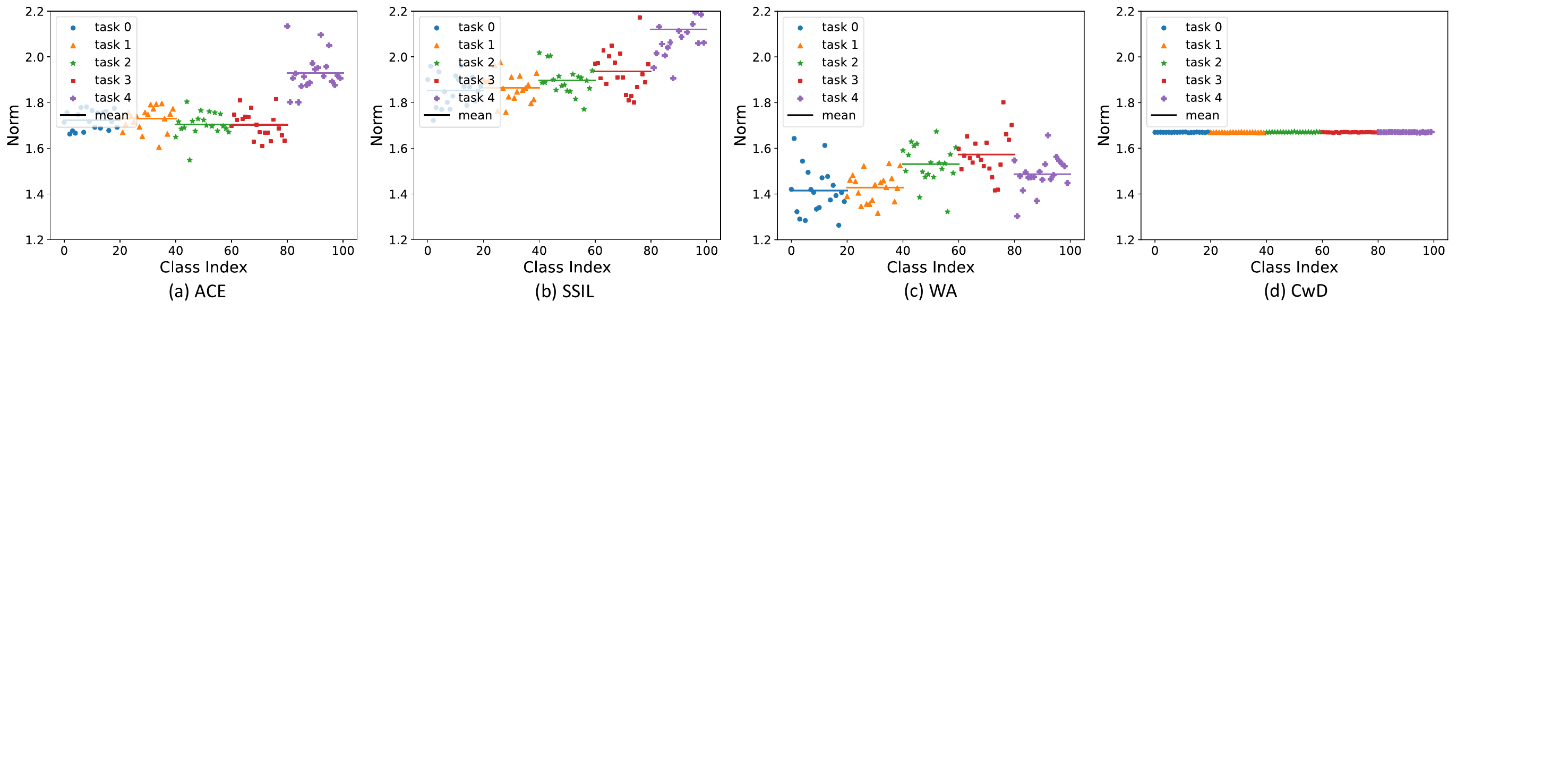}
  \caption{The norms of class weights in 5-task experiments with different debiasing approaches. (a) Asymmetric Cross-entropy. (b) Separated-Softmax for Incremental Learning. (c) Weight Aligning. (d) CwD with Weight Alignment Regularization.}
  \label{fig:debias}
\end{figure*}

We compare these methods by both the weight alignment situation and the final classification performance. The experiments are carried out on the CIFAR-100 dataset and the classification performance is reported in \Cref{tab:debias}. In addition, we observe the weight alignment situation at the end of the last task in the 5-task setting, as shown in \Cref{fig:debias}. Note that SCE performs similarly to ACE and the 2-task situation of SCE can be obtained by referring to \Cref{fig:norm}(d). 

From the results, we can see that SCE and ACE can mitigate the weight bias among old classes but can not mitigate the bias between new and old classes due to the data reason detailed in \Cref{sec:bias}. What is more, lacking information on non-target classes is attributed to the large drop in performance. SSIL intends to separate classes by tasks and keep information on non-target classes when distillation. Thus, though existing bias like SCE, it would not affect the performance much. As for WA, it performs the debiasing between old and new classes but lacks constraint within old classes, where the bias can be observed. What is more, the way of post-processing can not guarantee debiasing in the training stage which can result in a biased feature extractor and hurt the data consistency enhancement by contradicting the assumption. These factors incur the suboptimal performance of WA. Unlike them, our CwD avoids these disadvantages and explicitly mitigates the bias. 

\subsection{Ablation Study}
\begin{table}[t]
\centering
\caption{Ablation Study. We ablate each component of CwD. All numbers in the table are last incremental accuracy (mean$\pm$std\%).}
\begin{tabular}{lccc}
\hline
Method     & 5                & 10               & 20               \\ \hline
R-DFCIL    & 49.90 $\pm$ 0.23 & 42.57 $\pm$ 0.71 & 30.35 $\pm$ 0.12 \\
CwD$-$DCE & 50.95 $\pm$ 0.18 & 42.96 $\pm$ 0.98 & 31.08 $\pm$ 0.32 \\
CwD$-$WAR & 50.86 $\pm$ 0.34 & 43.52 $\pm$ 0.99 & 31.27 $\pm$ 0.19 \\
CwD     & \textbf{52.46 $\pm$ 0.35} & \textbf{43.69 $\pm$ 0.57} & \textbf{31.72 $\pm$ 0.22} \\ \hline
\end{tabular}
\label{tab:ablation}
\end{table}

The two main contributions of CwD are the proposed data consistency-enhanced (DCE) loss and the weight alignment regularization (WAR) loss. The DCE loss is applied in the inversion stage and the WAR loss is used in the training stage. To validate the effectiveness of every component of the proposed CwD, we ablate DCE and WAR respectively. As a result, we have 4 combinations: (1) CwD. The complete framework with both loss terms, (2) CwD$-$WAR. CwD framework without the WAR loss in the training stage, (3) CwD$-$DCE. CwD framework without DCE loss in the inversion stage, (4) R-DFCIL. CwD degrades to R-DFCIL without both DCE and WAR loss. We conduct experiments on CIFAR-100 as shown in \Cref{tab:ablation}.

Results suggest that each component can help improve the final incremental accuracy in all settings with different numbers of tasks. To be noticed, the DCE component has a more significant effect on the final performance compared to the WAR component. Further combining them achieves the best results.

\subsection{Parameter Analysis}
\begin{figure}[t]
  \centering
  \includegraphics[width=0.995\linewidth]{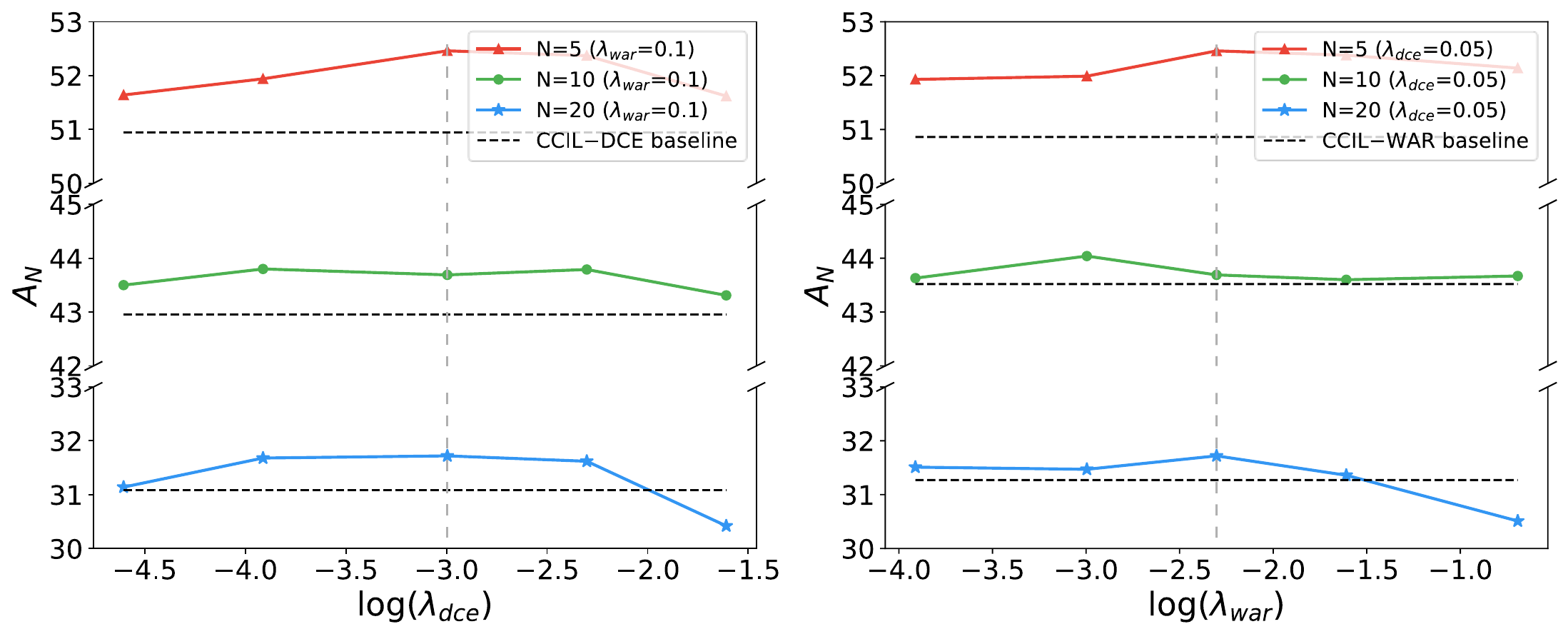}
  \caption{The influence of $\lambda_{dce}$ and $\lambda_{war}$ on CIFAR-100 with 5, 10 and 20 tasks. Left, search $\lambda_{dce}$ when $\lambda_{war}$ is fixed. Right, search $\lambda_{war}$ when $\lambda_{dce}$ is fixed.}
  \label{fig:parameters}
\end{figure}

In the CwD framework, we keep the hyperparameters the same as them in \cite{gao2022r} and there left two new hyperparameters $\lambda_{dce}$ and $\lambda_{war}$ to tune. We conduct the parameter analysis on CIFAR-100 with 5, 10, and 20 tasks. Because $\lambda_{dce}$ and $\lambda_{war}$ are introduced in the different stages, we perform two separate searches. First, we fix $\lambda_{war}=0.1$ and search the optimal $\lambda_{dce}$. We tried $\lambda_{dce}=0.01, 0.02, 0.05, 0.1, 0.2$. As shown in the left subfigure of \Cref{fig:parameters}, the last incremental accuracy first increases and then decreases as $\lambda_{dce}$ increases, which indicates that a local optimum exists. To be noticed, CwD performs better than the corresponding baseline in most cases. Then we fixed $\lambda_{dce}=0.05$ and search $\lambda_{war}$ in the range of $\{0.02, 0.05, 0.1, 0.2, 0.5\}$. Results in \Cref{fig:parameters} demonstrate that the performance is relatively less sensitive to $\lambda_{war}$ than $\lambda_{dce}$. The performance gain is close except when the regulation strength is set too high and the number of tasks is 20. Similarly, a wide range of values of $\lambda_{war}$ has positive effects on the final performance. According to the results, we set $\lambda_{dce}=0.05, \lambda_{war}=0.1$ to fit CIFAR-100 with different numbers of tasks and further adopt them in Tiny-ImageNet and ImageNet-100 experiments. More surprisingly, we also adopt the same parameters in the CwD + ABD experiments and it works well. Results in \Cref{sec:performance} validate the good transferability of the parameters.

{\color{black}\subsection{Quantitative Measurement of Data Consistency} 
\begin{table}[t]
\centering
\caption{Effects of Different Losses on Data Consistency. We ablate loss components of the baseline loss and show the improvement after combining $L_{dce}$.}
\begin{tabular}{lcc}
\hline
Method                                      & $D_{KL}$(KDE)         & $D_{KL}$(Gaussian)  \\ \hline
$L_{ce}$+$L_{div}$                          & 42.6510               & 253.9446         \\
$L_{stat}$+$L_{div}$                        & 33.2941               & 21.0037          \\
$L_{ce}$+$L_{stat}$+$L_{div}$               & 32.5018               & 22.5027          \\
$L_{ce}$+$L_{stat}$+$L_{div}$+$L_{dce}$     & \textbf{30.4094}      & \textbf{16.4198} \\ \hline
\end{tabular}
\label{tab:dkl}
\end{table}

In this section, we give the quantitative measurement of data consistency with respect to different loss combinations. We adopt both parametric estimation and non-parametric estimation to estimate $p(\boldsymbol{z})$ and $q(\boldsymbol{z})$ for a comprehensive understanding. For parametric estimation, we use the modeling in \Cref{sec:estimate}, while for non-parametric estimation, we employ the kernel density estimation (KDE) technique. Specifically, we use the Gaussian kernel and decide the bandwidth by Scott's method \cite{scott2015multivariate}. With estimated $p(\boldsymbol{z})$ and $q(\boldsymbol{z})$, one can calculate $D_{KL}$. But owing to no analytical solution and high computational complexity, we approximate KL divergence with Monte-Carlo approximation\footnote{{\color{black}We follow the algorithm in http://joschu.net/blog/kl-approx.html.}}. With the two KL Divergence metrics, we try to study the effects of loss combinations proposed in the literature under a unified framework. We first train a model with 50 classes in CIFAR-100 and then invert samples from it. Then, the KL divergences of inverted and real old data approximated in two ways are given. We ablate the loss components and the results can be found in \Cref{tab:dkl}. 

We can draw some conclusions from \Cref{tab:dkl}. First, the results of $D_{KL}$ (KDE) and $D_{KL}$ (Gaussian) are similar, which corroborate each other and can reflect the relative data consistency among methods. Second, the cross-entropy loss does not contribute to data consistency much. It is intuitive because cross-entropy loss is a sample-wise measurement and lacks any direct constraint on either the diversity or the distribution. What is more, minimizing the cross-entropy loss in \Cref{eq:ce} equals maximizing the Softmax score of samples. But it is well-studied that Softmax score~\cite{hendrycks2016baseline} is not a good metric for OOD detection~\cite{liu2020energy,morteza2022provable} and thus for data consistency measurement. Note in our experiments, the cross-entropy is not indispensable. We keep it for it does not contradict other losses during training and can slightly improve the performance in some cases. Third, the statistics alignment loss indeed ameliorates the consistency, but there is still plenty of room. We can observe that the alignment in \Cref{eq:stat} only considers the features of the BN layers but lacks explicit regularization on non-BN layer features. As a result, the proposed data consistency enhancement loss $L_{dce}$ on the features of the penultimate layer further enhance consistency as desired.}

{\color{black}\subsection{Analysis of Weight Bias} 
\label{sec:bias}

\begin{figure*}[t]
\centering
\includegraphics[width=0.95\linewidth]{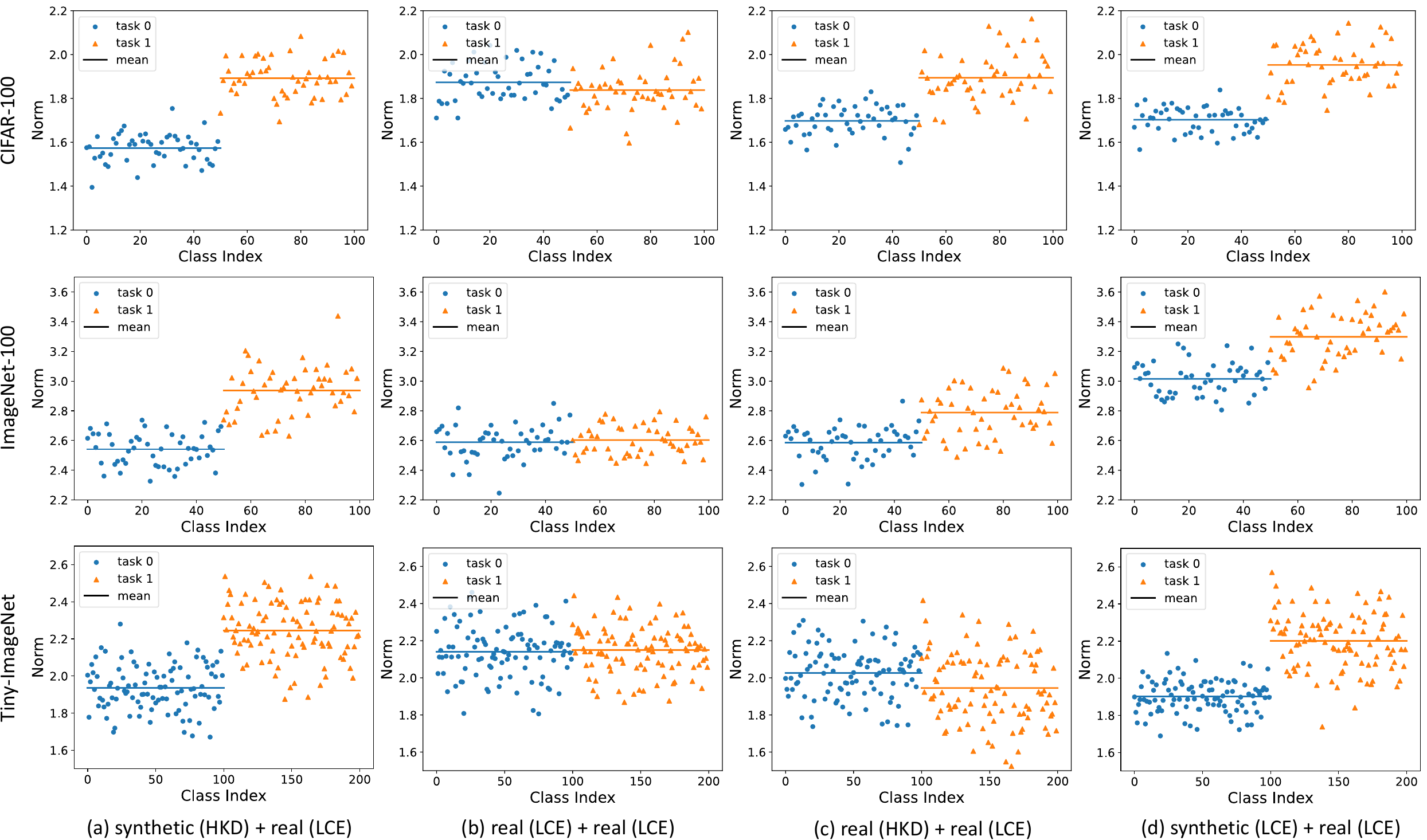}
{\color{black}\caption{The norms of class weights in 2-task experiments with different training schemes for the second task on CIFAR-100, ImageNet-100, and Tiny-ImageNet. (a) the standard training scheme. (b) a scheme where real old data is used and trained with local CE loss. (c) a scheme where the real old data is used. (d) a scheme where the synthetic old data is trained with the local CE loss. We report the results at the end of the final task.}
\label{fig:norm}}
\end{figure*}

A direct reason behind the class weight bias is the data imbalance problem. A similar observation is reported in \cite{zhao2020maintaining}, where data imbalance is caused by limited data memory. Different from that, the inversion-based methods can generate arbitrarily balanced and enough training samples. However, it is not the first choice in practice for the reason of low generation efficiency and poor generation quality. As a replacement, a batch of real new samples and a batch of generated old samples are used for training every iteration. The batch size is fixed across tasks. This will inevitably cause a data imbalance problem. For synthetic samples, the past class samples become fewer in a batch when past classes become more. But for real samples, the number of samples in each class is stable. What is more, the bias will increase as the incremental learning process continues due to the accumulation effect. 

However, when we control the number of classes of synthetic and real data to be the same, the weight bias problem still exists (as shown in \Cref{fig:norm}(a)). It indicates the existence of other reasons. We argue different loss functions and different training data may also contribute to weight bias and thus design a two-task class incremental experiment to uncover the factors concerned with it. The first task includes data from first 50\% classes while the second task includes the other 50\% classes. We conduct experiments on CIFAR-100, ImageNet-100, and Tiny-ImageNet. Training with standard CE is adopted for the first task. As for the second task, we compare 4 training schemes: (1) synthetic (HKD) - real (LCE), which is the standard training scheme, (2) real (LCE) - real (LCE), where real old data is used and trained with local CE loss, (3) real (HKD) - real (LCE), where the real old data is used, (4) synthetic (LCE) - real (LCE), where the synthetic old data is trained with the local CE loss. We report the norm of class weights at the end of the second task.

First of all, \Cref{fig:norm}(b) shows an unbiased situation under the condition of equal losses and data quality for both tasks. Further, results in \Cref{fig:norm}(c) and (d) validate our hypothesis that both the different training losses and different properties of training data will have effects on the class weights. HKD loss calculates the mean square error (MAE) of the logits of the old and new models as defined in \Cref{eq:hkd}. LCE loss defined in \Cref{eq:lce} is a cross-entropy loss applied locally among part of classes. To be noticed, MAE loss tends to oscillate near the optimal point while cross-entropy loss can always be minimized by larger class weight norms (\eg, by scaling). Thus, it is natural to find the norms of new class weights with LCE loss are bigger than those of old class weights with HKD loss in \Cref{fig:norm}(c) on CIFAR-100 and ImageNet-100 datasets. However, the situation changes on the Tiny-ImageNet dataset. We attribute it to the low classification accuracy, which indicates that the gradient direction of class weight vectors is not consistent. As a result, the norm will not increase consistently. In contrast, the gradient of HKD loss is more stable, resulting in larger norms.

\begin{figure}
  \centering
  \hspace{-0.5cm}\includegraphics[width=0.85\linewidth]{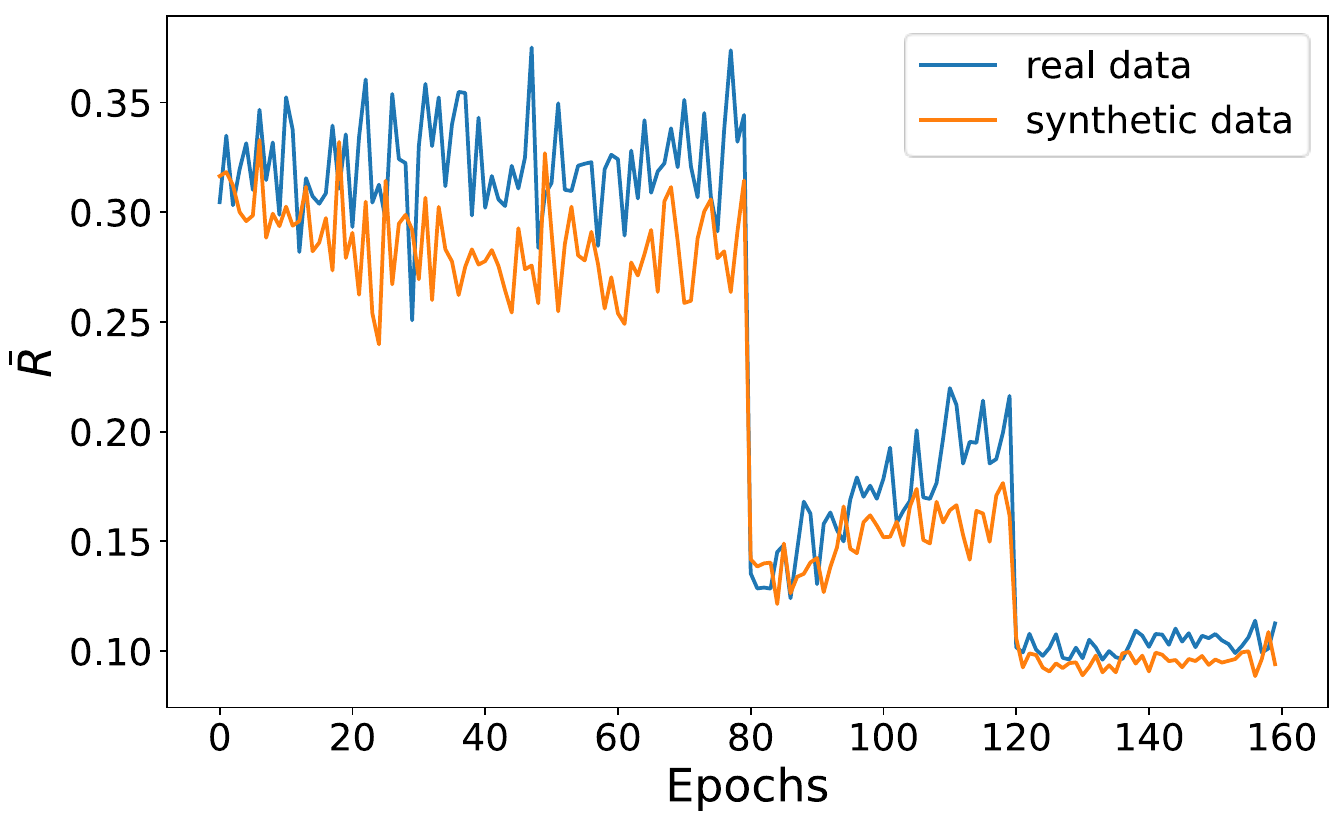}
  \caption{$\bar{R}$ of class weights belonging to synthetic and real data in the training process of task two in the synthetic (LCE) - real (LCE) scheme. $\bar{R}_{real}$ is larger than $\bar{R}_{synthetic}$, which indicates a more stable gradient direction of class weights of real data.}
  \label{fig:barR}
\end{figure}

On the other hand, we attribute the bias caused by synthetic samples (as shown in \Cref{fig:norm}(d)) to their poor separability. Due to the limited generation quality, the synthetic sample tends to mix many features of different classes~\cite{gao2022r}. Thus it introduces noisy information when training the classifier $g(\cdot)$ and resultantly causes a large variance in the gradient direction of $g(\cdot)$ with different samples. Assume that there are $M$ batches within one epoch. At each batch $B^m$, the derivative of $\mathcal{L}_{lce}^m$ with respect to $\boldsymbol{w_k}$ is $\frac{\partial \mathcal{L}_{lce}^m}{\partial \boldsymbol{w_k}}$. The gradient direction variance is reflected in the ratio of the norm of the mean gradient vector to the mean norm of gradient vectors, denoted as $R_k$ as follows:
\begin{equation}
R_k = \frac{||\frac{1}{M}\sum_{m=1}^M \frac{\partial \mathcal{L}_{lce}^m}{\partial \boldsymbol{w_k}}||}{\frac{1}{M}\sum_{m=1}^M||\frac{\partial \mathcal{L}_{lce}^m}{\partial \boldsymbol{w_k}}||}=\frac{||\sum_{m=1}^M \frac{\partial \mathcal{L}_{lce}^m}{\partial \boldsymbol{w_k}}||}{\sum_{m=1}^M||\frac{\partial \mathcal{L}_{lce}^m}{\partial \boldsymbol{w_k}}||}\leq 1.
\end{equation}
If the direction variance is large, $R_k$ tends to be small and vice versa. On top of that, we employ $\bar{R} = \frac{1}{|C|}\sum_{k\in C} R_k$ as the metric for measuring the variance regarding the norms of all class vectors. Notably, we split the dataset into batches to simulate the real scene during training. We plot $\bar{R}$ of classes belonging to synthetic and real samples in the synthetic (LCE) - real (LCE) scheme as shown in \Cref{fig:barR}. It shows that the gradient direction variance of class weights of real data is smaller than that of synthetic data in the whole training stage. It is conducive to a fast convergence and stable promotion of the weight norm, which explains the phenomenon of \Cref{fig:norm}(d). The new observation about data quality suggests explicit debiasing strategy is preferred and debiasing methods \cite{ahn2021ss,caccia2022new} focusing on split loss functions may not be compatible with synthetic data in DFCIL. Results of the comparison with different debiasing methods can be found in \Cref{sec:debias}.}

{\color{black}\subsection{Computational Overhead}
\label{sec:overhead}
Due to the existence of the estimation stage, it needs more computation than the baseline method. In this section, we report the exact computation overhead. Denote the forward-pass overhead of a data batch through the classification model as $c_p$, the backward-pass overhead as $c_b$, the forward-pass overhead of a noise batch through the generator as $g_p$, the backward-pass overhead as $g_b$, the number of iterations of the inversion stage is $N_{inv}$, the number of epochs in the training stage is $N_{e}$, and the number of batches in one epoch is $M$. Then, we can obtain the computational overhead for the inversion stage, training stage, and estimation stage as $N_{inv}\times(g_p+g_b+c_p+c_b)$, $N_{e}\times M\times(g_p+2\times(c_p+c_b))$, and $M\times(|\mathcal{C}_{old}|/|\mathcal{C}_{new}|\times(g_p+c_p) +c_p)$. In the settings of our experiments, $|\mathcal{C}_{old}|/|\mathcal{C}_{new}|\ll N_e$ and $M\times|\mathcal{C}_{old}|/|\mathcal{C}_{new}|\ll N_{inv}$. As a result, the extra computational overhead of the estimation stage is negligible. We show the practical running time in \Cref{tab:cost}, which is consistent with the theoretical complexity.}

\begin{table*}[t]
\scriptsize
\centering
{\color{black}\caption{Computational overhead on different benchmarks. $V_{a}^b$ denotes the cost for ER with $a$ tasks on $b$ dataset.}
\label{tab:cost}}
\arrayrulecolor{black}
\setlength{\tabcolsep}{1.6mm}{\begin{tabular}{lccccccccc}
\hline
           & \multicolumn{3}{c}{\textcolor{black}{CIFAR-100}}                          & \multicolumn{3}{c}{\textcolor{black}{Tiny-ImageNet}}                    & \multicolumn{3}{c}{\textcolor{black}{ImageNet-100}}                   \\ \cline{2-10} 
           & \textcolor{black}{5}               & \textcolor{black}{10}               & \textcolor{black}{20}                & \textcolor{black}{5}              & \textcolor{black}{10}              & \textcolor{black}{20}                & \textcolor{black}{5}              & \textcolor{black}{10}             & \textcolor{black}{20}               \\ \hline
\textcolor{black}{ER}         & \textcolor{black}{$V_5^C$}         & \textcolor{black}{$V_{10}^C$}       & \textcolor{black}{$V_{20}^C$}        & \textcolor{black}{$V_5^T$}        & \textcolor{black}{$V_{10}^T$}      & \textcolor{black}{$V_{20}^T$}        & \textcolor{black}{$V_5^I$}        & \textcolor{black}{$V_{10}^I$}     & \textcolor{black}{$V_{20}^I$}       \\ 
\textcolor{black}{R-DFCIL}    & \textcolor{black}{1.14$V_5^C$}     & \textcolor{black}{1.35$V_{10}^C$}   & \textcolor{black}{1.77$V_{20}^C$}    & \textcolor{black}{1.12$V_5^T$}    & \textcolor{black}{1.62$V_{10}^T$}  & \textcolor{black}{1.91$V_{20}^T$}    & \textcolor{black}{1.03$V_5^I$}    & \textcolor{black}{1.21$V_{10}^I$} & \textcolor{black}{1.46$V_{20}^I$}   \\ 
\textcolor{black}{CwD}        & \textcolor{black}{1.17$V_5^C$}     & \textcolor{black}{1.47$V_{10}^C$}   & \textcolor{black}{2.04$V_{20}^C$}    & \textcolor{black}{1.15$V_5^T$}    & \textcolor{black}{1.68$V_{10}^T$}  & \textcolor{black}{2.03$V_{20}^T$}    & \textcolor{black}{1.05$V_5^I$}    & \textcolor{black}{1.24$V_{10}^I$} & \textcolor{black}{1.52$V_{20}^I$}   \\ \hline
\end{tabular}}
\arrayrulecolor{black}
\end{table*}

{\color{black}\section{Limitations}
\label{sec:lim}
One limitation of CwD is the increased computational cost compared to non-data-free baselines like ER. The introduction of both the inversion stage and the estimation stage in each task results in higher computational demands. Detailed information on the computational overhead is provided in \Cref{sec:overhead}. Another potential limitation is the restricted modeling of the data distribution. Currently, our modeling approach is limited to the features of the penultimate layer under a multivariate Gaussian assumption. However, statistics from other layers may also contribute to addressing the inversion challenge. Exploring and incorporating these additional statistics is a promising direction for future research and may lead to improved performance. Lastly, by applying $\mathcal{L}_{war}$ in the training stage, CwD does not preserve the relative magnitudes of weight vector norms within the task. We hypothesize that task-wise alignment, rather than class-wise alignment, may suffice to mitigate bias in CwD, which we plan to investigate in future work.}

\section{Conclusion}
\label{sec:Conclusion}
In this paper, we propose CwD framework to solve the catastrophic forgetting problem in data-free incremental learning. To address the data inconsistency problem in the literature, we first propose the quantitative measure of data consistency, which further inspires the development of a novel loss term. Specifically, by aligning the statistical parameters in the feature space, we narrow the gap between synthetic and real data and ameliorate the inversion stage. This approach proves to be easy to implement. Furthermore, we identify a phenomenon where the norms of old class weights decrease as learning progresses. We analyze the underlying reasons in the background of DFCIL and propose a simple and effective regularization term to reduce the weight bias. Experiments on different datasets show our method can surpass previous works and achieve SOTA performance. Also importantly, we believe the comprehensive analysis of multiple aspects of data-free data replay methods in our study will contribute to the ongoing efforts to develop more effective techniques.

\section*{Acknowledgement}
The research was supported by the National Natural Science Foundation of China (U23B2009, 92270116).



\bibliographystyle{elsarticle-num-names} 
\bibliography{ccil}




\end{document}